\pdfoutput=1 
\documentclass[11pt,a4paper]{article}
\usepackage[hyperref]{emnlp-ijcnlp-2019}
\usepackage{times}
\usepackage{latexsym}
\usepackage{amssymb}
\usepackage[inline]{enumitem}
\setlist[enumerate,1]{label=\textit{\alph*)}}
\usepackage[utf8]{inputenc}
\usepackage[T1]{fontenc}
\usepackage[spanish,english]{babel}
\usepackage{url}
\usepackage{multirow}
\usepackage{xcolor}
\newcommand{\ensuretext}[1]{#1}
\newcommand{\mcmarker}{\ensuretext{\textcolor{magenta}{\ensuremath{^{\textsc{M}}_{\textsc{C}}}}}}

\newcommand{\mycomment}[3]{\ensuretext{\textcolor{#3}{[#1 #2]}}}
\newcommand{\mc}[1]{\mycomment{\mcmarker}{#1}{magenta}}

\newcommand{\ignore}[1]{}
\usepackage{graphicx}
\usepackage{booktabs}
\usepackage{pifont}
\newcommand{\cmark}{\ding{51}}%
\newcommand{\loss}{\mathcal{L}}

\aclfinalcopy 

\title{Controlling Text Complexity in Neural Machine Translation}

\author{Sweta Agrawal \\
  Department of Computer Science \\
  University of Maryland \\
  {\tt sweagraw@cs.umd.edu} \\\And
  Marine Carpuat \\
  Department of Computer Science \\
  University of Maryland \\
  {\tt marine@cs.umd.edu} \\}

\date{}

\begin{document}
\maketitle
\begin{abstract}

This work introduces a machine translation task where the output is aimed at audiences of different levels of target language proficiency. We collect a high quality dataset of news articles available in English and Spanish, written for diverse grade levels and propose a method to align segments across comparable bilingual articles. The resulting dataset makes it possible to train multi-task sequence-to-sequence models that translate Spanish into English targeted at an easier reading grade level than the original Spanish. We show that these multi-task models outperform pipeline approaches that translate and simplify text independently.

\end{abstract}



\section{Introduction}


Generating text at the right level of complexity can make machine translation (MT) more useful for a wide range of users. As \newcite{XuCallison-BurchNapoles2015} note, simplifying text makes it possible to develop reading aids for people with low-literacy \cite{WillianWatanabeArnaldJunioro2009, JanDeBelderMarie-FrancineMoens2010}, for non-native speakers and language learners \cite{SarahPetersenMariOstendorf2007, DavidAllen2009}, for people who suffer from language impairments \cite{JohnCarrollGuidoMinnenDarrenPearce1999, YvonneCanningJohnTaitJackieArchibaldRosCrawley2000, KentaroInuiAtsushiFujitaTetsuroTakahashiRyuIida2003}, and for readers lacking expert knowledge of the topic discussed \cite{NoemieElhadadKomalSutaria2007,AdvaithSiddharthanNapoleonKatsos2010}. Such readers would also benefit from MT output that is better targeted to their needs by being easier to read than the original. 


Prior work on text complexity has focused on simplifying input text in one language, primarily English \cite{ChandrasekarDoranSrinivas1996,CosterKauchak2011,Siddharthan2014,XuCallison-BurchNapoles2015,ZhangLapata2017, ScartonSpecia2018, KrizSedocApidianakiZhengKumarMiltsakakiCallsion-Burch2019, NishiharaKajiwaraArase2019}. 
Simplification has been used to improve MT by restructuring complex sentences into shorter and simpler segments that are easier to translate \cite{GerberHovy1998,SanjaStajnerMajaPopovic2016, HaslerdeGispertStahlbergWaiteByrne2017}. 
Contemporaneously to our work, \citet{MarchisioGuoLaiKoehn2019} show that tagging the English side of parallel corpora with automatic readability scores can help translate the same input in a simpler or more complex form. Our work shares the goal of controlling translation complexity, but considers a broader range of reading grade levels and simplification operations grounded in professionally edited text simplification corpora.

Building a model for this task ideally requires rich annotation for evaluation and supervised training that is not available in bilingual parallel corpora typically used in MT. Controlling the complexity of Spanish-English translation ideally requires examples of Spanish sentences paired with several English translations that span a range of complexity levels. We collect such a dataset of English-Spanish segment pairs from the Newsela website, which provides professionally edited simplifications and translations. By contrast with MT parallel corpora, the English and Spanish translations at different grade levels are only comparable. They differ in length and sentence structure, reflecting complex syntactic and lexical simplification operations.
 
We adopt a multi-task approach to control complexity in neural MT and evaluate it on complexity-controlled Spanish-English translation. 
Our extensive empirical study shows that multitask models produce better and simpler translations than pipelines of independent translation and simplification models. We then analyze the strengths and weaknesses of multitask models, focusing on the degree to which they match the target complexity, and the impact of training data types and reading grade level annotation.\footnote{Researchers can request the bilingual Newsela data at \url{https://Newsela.com/data/}. Scripts to replicate our model configurations and our cross-lingual segment aligner are available at \url{https://github.com/sweta20/ComplexityControlledMT}.}  



\section{Background}

Given corpora of parallel complex-simple segments, text simplification can naturally be framed as a translation task, borrowing and adapting model architectures originally designed for MT. \citet{XuNapolesPavlickChenCallison-Burch2016} provide a thorough study of statistical MT techniques for English text simplification, and introduce novel objectives to measure simplification quality. Interestingly, they indirectly make use of parallel translation corpora to derive simplification paraphrasing rules by bilingual pivoting \cite{Callison-Burch2007}. \citet{ZhangLapata2017} train sequence-to-sequence models to translate from complex to simple English using reinforcement learning to directly optimize the metrics that evaluate complexity (SARI) and fluency and adequacy (BLEU). \citet{ScartonSpecia2018} address the task of producing text of varying levels of complexity for different target audiences. They show that neural sequence-to-sequence models informed by target-complexity tokens inserted in the input sequence  perform well on this task. While the vast majority of text simplification work has focused on English, Spanish \citep{StajnerCalixtoSaggion2015}, Italian \citep{BrunatoCiminoDell'OrlettaVenturi2016, AprosioTonelliTurchiNegriDiGangi2019} and German \citep{KlaperEblingVolk2013} have also received some attention. 


While most MT approaches only indirectly capture style properties (e.g., via  domain adaptation), a growing number of studies share the goal of considering source texts and their translation in their pragmatic context.   \citet{MirkinMeunier2015} introduce personalized MT.  \citet{RabinovichMirkinPatelSpeciaWintner2016} and \citet{VanmassenhoveHardmeierWay2018} suggest that the gender of the author is implicitly marked in the source text and that dedicated statistical and neural systems  better preserve gender traits in MT output.  Neural MT has enabled more flexible ways to control  stylistic properties of MT output. \citet{SennrichHaddowBirch2016c} first propose to append a special token to the source that neural MT models can attend to and to select the formal (Sie) or informal (du) version of second person pronouns when translating into German. \citet{NiuRaoCarpuat2018} show that multi-task models can jointly translate between languages and styles, producing formal and informal translations with broader lexical and phrasal changes than the local pronoun changes in \citet{SennrichHaddowBirch2016c}. Closest to our goal, \citet{MarchisioGuoLaiKoehn2019} address the task of producing either simple or complex translations of the same input, using automatic readability scoring of parallel corpora. They show that training distinct decoders for simple and complex language allows better complexity control than using the target complexity as a side-constraint. By contrast, our approach exploits text simplification corpora for richer supervision for both training and evaluation.



\section{A Multi-Task Approach to Complexity Controlled MT}
\label{sec:model}

\paragraph{Task} We define \textbf{complexity controlled MT} as a task that takes two inputs: an input language segment $s_i$ and a target complexity $c$. The goal is to produce a translation $s_o$ in the output language that has complexity $c$. For instance, given input Spanish sentences in Table~\ref{tab:example1}, complexity controlled MT aims to produce English translations at a specific level of complexity, which might differ from the complexity of the original Spanish. 

\paragraph{Model} We model $P(s_o|s_i,c;\theta)$ as a neural encoder-decoder with attention \cite{BahdanauChoBengio2015}. This architecture has been used successfully for the two related tasks of text simplification  \cite{ WangChenRochfordQiang2016,ZhangLapata2017,SergiuNisioiSanjaStajnerPaoloPonzettoLiviuDinu2014,ScartonSpecia2018} and machine translation \cite{BahdanauChoBengio2015}. The encoder constructs hidden representation for each word in the input sequence, while the decoder generates the target sequence, conditioned on hidden source representations. 
We hypothesize that training a single encoder-decoder model to perform the two distinct tasks of machine translation and text simplification will yield a model that can perform complexity controlled MT. We adopt the multi-task framework proposed by  \newcite{JohnsonSchusterLeKrikunWuChenThoratViegasWattenbergCorradoHughesDean2016} to train multilingual neural MT systems. 

\begin{table*}[ht]
\centering
\begin{tabular}{lp{5.5cm}lp{5.5cm}p{2cm}}
 \toprule
 $c_i$ & Spanish ($s_i$) & $c_o$ &  English ($s_o$) & Operation\\
 \midrule
 9 & Doug Ratliff, un empresario de 67 años de edad de Richlands, Virginia, dijo que la elección de Trump sería uno de los días más felices de su vida. &  3 & Doug Ratliff is a businessman from Virginia. Ratliff said Trump's election would be one of the happiest days of his life. & Splitting; Deletion \\
    \midrule
12 & Incluso antes de haber nacido, Daliyah Marie Arana, según dicen sus padres, estaba aprendiendo a leer. &4 &Daliyah Marie Arana has been learning to read since before she was born. & Paraphrasing\\
\midrule
9  & Kes Gray es el escritor de la serie de cuentos de animales "Oi Frog and Friends". A él no le interesaron mucho los descubrimientos del estudio. Los autores y los ilustradores solo necesitan que los personajes principales animales de sus historias sean adorables, concluyó. & 5 & Kes Gray is the writer of the rhyming animal series "Oi Frog and Friends." He was not bothered by the study’s findings. Writers and artists just need to keep the main animal characters in their stories cuddly, he said. & Lexical substitution\\
\bottomrule
\end{tabular}
\caption{Cross-lingual Newsela examples: the Spanish text $s_i$ of complexity, or reading grade level, $c_i$ is automatically aligned to English text $s_o$ of $c_o$. Simplification transformations range from sentence splitting and deletions to paraphrasing and lexical substitution.}\label{tab:example1}
\end{table*}

\paragraph{Representing target complexity} Target complexity $c$ can be incorporated in sequence-to-sequence models as a special token appended to the beginning of the input sequence, which acts as a side constraint. The encoder encodes this token in its hidden states as any other vocabulary token, and the decoder can attend to this representation to guide the generation of the output sequence. This simple strategy has been used in MT to control second person pronoun forms when translating into German \cite{SennrichHaddowBirch2016c}, to indicate the target language in multilingual MT \cite{JohnsonSchusterLeKrikunWuChenThoratViegasWattenbergCorradoHughesDean2016}, and to obtain formal or informal translations of the same input \cite{NiuRaoCarpuat2018}. In monolingual text simplification tasks \cite{ScartonSpecia2018}, the reading grade level has been encoded as such a special token. 

\paragraph{Training Data and Objectives} 
Fully supervised training would ideally require translation samples with outputs representing different levels of complexity for the same input segment. However, constructing such data at the scale required to train deep neural networks is expensive and unrealistic. 
Our multi-task training configuration lets us exploit different types of training examples to train shared encoder-decoder parameters $\theta$.
We use the following samples/tasks:
\begin{itemize}
	\item Complexity controlled MT samples  $(s_i,c_o,s_o)$:  These are the closest samples to the task at hand, but are hard to obtain. They are used to defined the complexity-controlled MT loss
	\begin{equation}\label{eq:loss}
\loss_{CMT} = \sum_{(s_i,c_o,s_o)} \log P( s_o |  s_i , c_o  ; \theta)
\end{equation}
	\item MT samples $(s_i,s_o)$: These are sentence pairs drawn from parallel corpora. They are available in large quantities for many language pairs \cite{Tiedemann2012} and are used to define the MT loss
	\begin{equation}\label{eq:loss}
\loss_{MT} = \sum_{(s_i,s_o)} \log P( s_o |  s_i  ; \theta)
\end{equation}
	\item Text simplification samples in the MT target language $(s_o,c_{s_o'},s_o')$ where $s_o'$ is a simplified version of complexity $c_{s_o'}$ for input $s_o$, which are likely to be available in much smaller quantities than MT samples.
	\begin{equation}\label{eq:loss}
\loss_{\textit{Simplify}} = \sum_{(s_o,c_{s_o'},s_{o}')} \log P( s_o' |  s_o, c_{s_o'}  ; \theta)
\end{equation}
\end{itemize}
The multi-task loss is simply obtained by summing the losses from individual tasks: $\loss_{CMT} + \loss_{MT}+\loss_{Simplify}$.

\section{The Newsela Cross-Lingual Simplification Dataset}
\label{NewselaESEN}


We build on prior work that used the Newsela dataset for English or Spanish text simplification by automatically aligning English and Spanish segments of different complexity to enable complexity-controlled machine translation.

The Newsela website provides high quality data to study text simplification.  \citet{XuCallison-BurchNapoles2015} argue that text simplification research should be grounded in texts that are simplified by professional editors for specific target audiences, rather than more general-purpose crowd-sourced simplifications such as those available on Wikipedia. They show that Wikipedia is prone to sentence alignment errors, contains a non-negligible amount of inadequate simplifications, and does not generalize well to other text genres. By contrast, Newsela is an instructional content platform meant to help teachers prepare curriculum that match the language skills required at each grade level. The Newsela corpus consists of English articles in their original form, 4 or 5 different versions rewritten by professionals to suit different grade levels as well as optional translations of original and/or simplified English articles into Spanish resulting in 23,130 English and 5,320 Spanish articles with grade annotations respectively.

This section introduces our method to align English and Spanish segments across complexity levels, and the resulting bilingual dataset.

\subsection{Cross-Lingual Segment Alignment}


Extracting training examples from this corpus requires aligning segments within documents. This is challenging because text is neither simplified nor translated sentence by sentence, and as a result, equivalent content might move from one sentence to the next. 
Past work has introduced techniques to align segments of different complexity within documents of the same language \citep{XuCallison-BurchNapoles2015,PaetzoldManchegoSpecia2017,StajnerSanjaMarcPaoloSimone2018}.

Complexity controlled MT requires aligning segments of different complexity in English and Spanish. 
Existing methods for aligning sentences in English and Spanish parallel corpora are not well suited to this task. For instance, the Gale-Church algorithm \cite{GaleChurch1993} assumes that aligned sentences should have similar length. This assumption does not hold if the English article is a simplification of the Spanish article.  Consider the following Spanish text and its English translation in Newsela:

\begin{otherlanguage}{spanish}
    \textbf{Spanish}: LA HAYA, Holanda - Te has tomado alguna vez una selfie?, Hoy en día es muy fácil. Solo necesitas un teléfono inteligente.
    
    \textbf{Google Translated English}: THE HAGUE, Netherlands - Have you ever taken a selfie? Today is very easy. You only need a smart phone. 
    
    \textbf{Original English Version}: THE HAGUE, Netherlands - All you need is a smartphone to take a selfie. It is that easy.
\end{otherlanguage}

As a result, we adapt a monolingual text simplification aligner for cross-lingual alignment. MASSAlign \cite{PaetzoldManchegoSpecia2017} is a Python library designed to align segments of different length within comparable corpora of the same language. It employs a vicinity-driven search approach,  based on the assumption that the order in which information appears is roughly constant in simple and complex texts. A similarity matrix is created between the paragraphs/sentences of aligned documents/paragraphs using a standard bag-of-words TF-IDF model. It finds a starting point to begin the search for an alignment path, allowing long-distance alignment skips, capturing 1-N and N-1 alignments. To leverage this alignment flexibility, we apply MASSAlign  to English articles and Spanish articles machine translated into English by Google translate.\footnote{\url{https://translate.google.com/}} An important property of Google translated articles is that they are aligned 1-1 at the sentence level. This lets us deterministically find the Spanish replacement for the aligned Google translated English version returned by MASSAlign. Translation quality is high for this language pair, and even noisy translated articles contain enough signal to construct the similarity matrix required by MASSAlign.

\begin{table*}[ht]
\centering
\begin{tabular}{lrrrr}
 \toprule
 Dataset  & \# tokens/segment &  \#sents/segment  & \# of types & \# of tokens \\
 \midrule
 \textit{Spanish} \\
  ~~Newsela & 50.13 & 2.17& 57,361 & 7,792,285\\
 ~~Global Voices & 22.96 & 1.03 & 254,111 & 15,921,948\\
 ~~News & 26.73 & 1.03 &  80,840 & 5,587,307\\
  \textit{English} \\
  ~~Newsela & 43.37  & 2.65 & 39,012 & 7,139,717 \\
 ~~Global Voices &  21.93  &1.06 & 222,383 & 15,208,054\\
 ~~News & 23.76 &  1.04 & 49,589 & 4,939,085\\
  \bottomrule
\end{tabular}
\caption{Comparisons of the English-Spanish Newsela corpus with machine translation corpora from OPUS drawn from Global Voices and News Commentary. \label{tab:stats1} }
\end{table*}

\subsection{Resulting Dataset}

We thus create:  both  samples for complexity controlled MT $(s_i,c_o,s_o)$ and traditional monolingual text simplification samples $ \left( s_o,c_{s_o'},s_o'\right)$ that can be used by the multi-task model (Section~\ref{sec:model}). Since the properties of Newsela monolingual simplification samples have been studied thoroughly by \newcite{XuCallison-BurchNapoles2015}, we present key statistics for the cross-lingual simplification examples only. Table~\ref{tab:stats1} contrasts Newsela parallel segments with bilingual parallel sentences drawn from the OPUS corpus \citep{Tiedemann2009}. We use Global Voices and News Commentary from OPUS corpus as it has the most similar domain to the Newsela data. Aligned segments in Newsela are about twice as long as segments in parallel corpora, and contain more than two sentences on each side on average. By contrast, parallel corpora samples align sentences one-to-one on average. 

Articles are distributed across reading levels spanning grades 2 to 12 for both English and English-Spanish pairs.  Table~\ref{tab:stats} highlights the vocabulary differences among the different grade levels for the Newsela Spanish-English corpus. The vocabulary size of the corpus corresponding to lower grade level is smaller as compared to higher complexity levels. Also, complex sentences have more words per sentence on average but fewer sentences per segment compared to their simplified counterparts. Simple sentences differ from complex sentences in various ways, ranging from sentence splitting and content deletion to paraphrasing and lexical substitutions, as illustrated in Table~\ref{tab:example1}. 

\addtolength{\tabcolsep}{-1pt}    
\begin{table*}[ht]
\centering
\begin{tabular}{lrrrrrr}
 \toprule
   & \multicolumn{3}{c}{Source (Spanish)}&  \multicolumn{3}{c}{Target (English)} \\
 Grade & word types & tokens/segment &  sents/segment & word types & tokens/segment &  sents/segment \\
 \midrule
2 & - & - & - & 3749 & 34.31 & 3.76  \\
3 & 2,628 & 38.59 & 3.52 & 10,615 & 35.57 & 3.28 \\
4 & 8,431 & 39.33 & 2.95 & 10,414 & 39.38 & 3.09\\
5 & 17,082 & 40.96 & 2.59 & 18,508 & 42.18 & 2.87 \\
6 & 16,945 & 42.78 & 2.25 & 16,613 & 44.21 & 2.62\\
7 & 22,352 & 46.65 & 2.19 & 23,617 & 47.54 & 2.57\\
8 & 19,317 & 47.07 & 1.96 & 17,746 & 46.66 & 2.24 \\
9 & 24,846 & 50.08 & 1.87 & 22,230 & 50.08 & 2.25 \\
10 & 482 & 49.19 & 1.90 & 341 & 38.23 & 1.70\\
12 & 42,355 & 53.98 & 1.96 & - & - & -\\
  \bottomrule
\end{tabular}
\caption{Grade level Statistics of the Newsela Spanish-English corpus. Vocabulary size decreases with the reading grade level. Simpler segments contain fewer sentences and are often shorter that complex segments. 
\label{tab:stats}}
\end{table*}
\addtolength{\tabcolsep}{1pt}

\section{Experiment Settings}
\label{sec:setup}

We evaluate \textbf{complexity controlled MT} using a subset of the 150k Spanish-English segment pairs extracted from Newsela as described in Section~\ref{NewselaESEN}. We select Spanish and English segments that have different reading grade levels, so that given a Spanish input, the task consists in producing an English translation which is simpler (lower reading grade level) than the Spanish input.
 The train/development/test split ensures that there is no overlap between articles held out for testing and articles used for training. We refer to the corresponding training examples as  \textbf{MT+simplify} since it represents the joint task of translation and simplification. 

\subsection{Evaluation Metrics}



We evaluate the truecased detokenized output of our models using three automatic evaluation metrics, drawing from both machine translation and text simplification evaluation.

BLEU \cite{PapineniRoukosWardZhu2002} estimates translation quality based on $n$-gram overlap between system output and references. However it does not separate mismatches due to meaning errors and mismatches due to simplification errors.

SARI \cite{XuNapolesPavlickChenCallison-Burch2016}\footnote{\url{https://github.com/cocoxu/simplification}} is designed to evaluate text simplification systems by comparing system output against references and against the input sentence. It explicitly measures the goodness of words that are added, deleted and kept by the systems.  \newcite{XuNapolesPavlickChenCallison-Burch2016} showed that BLEU shows high correlation with human scores for grammaticality and meaning preservation and SARI shows high correlation with human scores for simplicity. In the cross-lingual setting, we cannot directly compare the Spanish input with English hypotheses and references, therefore we use the baseline machine translation of Spanish into English as a pseudo-source text. The resulting SARI score directly measures the improvement over baseline machine translation.

In addition to BLEU and SARI, we report Pearson's correlation coefficient (PCC) to measure the strength of the linear relationship between the complexity of our system outputs and the complexity of reference translations.  \citet{HeilmanCollinsEskenazi2008} use it to evaluate the performance of reading difficulty prediction. 
    Here we estimate the reading grade level complexity of MT outputs and reference translations using the Automatic Readability Index (ARI)\footnote{\url{https://github.com/mmautner/readability}} score, which combines evidence from the number of characters per word and number of words per sentence using hand-tuned weights \citep{SenterSmith1967}:
\begin{equation}
\label{ariequation}
 ARI = 4.71(\frac{chars}{words}) + 0.5( \frac{words}{sents}) - 21.43
\end{equation}

\ignore{
    \item Automatic Readability Index (ARI) \cite{SenterSmith1967} based adjacency accuracy \mc{bad name and misleading acronym. collocation too long. reviewer suggests: adjacency accuracy based on ARI. perhaps rename to something that more directly conveys that this measures whether the output achieves the target grade}. In \citet{HeilmanCollinsEskenazi2008}, adjacency accuracy was used for evaluating the performance of reading difficulty predictions. It is defined as the proportion of predictions that were within one grade level of the human assigned grade for the given text. 

\begin{equation}
\label{ariequation}
 \text{ARI}
 = 4.71(\frac{chars}{words}) + 0.5( \frac{words}{sents}) - 21.43
\end{equation}
\mc{Reviewer: any explanation for curious weights?}
}

\subsection{Training Data}

In addition to the Newsela \textbf{MT+Simplify} training examples described above, which are of the form $(s_i,c_o,s_o)$, we use monolingual English simplification data, bilingual parallel training data and Spanish simplification data.

\paragraph{Newsela Simplification}  provides training examples of the form $(s_o,c_{s_o'},s_o')$, where $s_o$ and $s_o'$ are in the same language. We refer to this data as \textbf{Simplify} data. It is used for training multi-task models and for auxiliary evaluation on English only. Our version of this corpus has 513k English segment pairs extracted using the method by \citet{PaetzoldSpecia2016}. Similar to \citet{ScartonSpecia2018}, an original article 0 can be aligned to up to four simplified versions: 1,2,3 and 4. Here 4 denotes the least simplified level and 0 represents the most simplified level. The train split consists of 460k instance pairs whereas the development and test sets consist of roughly 20K instances, drawn from the same articles as the MT+preserve and MT+simplify test set. For Spanish, we have  110k segment pairs, which will be used to train the Spanish simplification baseline.

\paragraph{Bilingual Parallel Data (Newsela)} We also extract parallel Spanish-English segments from Newsela based on aligned segments between Spanish and English articles that have the same reading grade level. We use this dataset to provide in-domain MT training examples which includes roughly 70k instances.

All datasets are pre-processed using Moses tools for normalization, tokenization and true-casing \citep{KoehnHoangBirchCallison-BurchFedericoBertoldiCowanShenMoranZensDyerBojarConstantinHerbst2007}. We further segment tokens into subwords using a joint source-target byte pair encoding model with 32,000 operations \cite{SennrichHaddowBirch2015}.

\subsection{Sequence-to-Sequence Model Configuration}
\label{modelConfig}

We use the standard encoder-decoder architecture implemented in the Sockeye toolkit \cite{Hieber2017}. Both encoder and decoder have two Long Short Term Memory (LSTM) layers \citep{BahdanauChoBengio2015}, hidden states of size 500 and dropout of 0.3 applied to the RNNs of the encoder and decoder which is same as what was used by \newcite{ScartonSpecia2018}. We observe that dot product based attention works best in our scenario, perhaps indicating that the task of complexity controlled translation requires mostly local changes that do not lead to long distance reorderings across sentences. We train using the Adam \citep{KingmaBa2014} optimizer with a batch size of 256 segments and checkpoint the model every 1000 updates. Training stops after 8 checkpoints without improvement of validation perplexity. The vocabulary size is limited to 50000. We decode with a beam size of 5. Grade side-constraints are defined using a distinct special token for each grade level (from 2 to 12). The constraint token corresponds to the grade level of the target instance.

\subsection{Baseline}
\label{pipeline}

We contrast the multi-task system with pipeline based approaches, where translation and simplification are treated as independent consecutive steps. We train a neural MT model to perform translation from Spanish to English and other neural MT models to perform monolingual text simplification for Spanish and English respectively. In the first pipeline setup, the output from the translation model is fed as input to an English simplification model while in the other, the output from the Spanish simplification model is fed as input to an translation model. As \newcite{ScartonSpecia2018}, we simply use grade level tokens as side constraints on English simplification examples to control output complexity.\footnote{Additional constraints based on simplification operations were also used in that work but did not provide substantial benefits when operations are predicted based on the input.}

\section{Evaluation of Complexity Controlled MT}\label{sec:results}



We compare pipeline and multitask models on the Newsela complexity controlled MT task (Table~\ref{tab:mainresult}). Overall, results show that compared to pipeline models, multitask models produce complexity controlled translations that better match human references according to BLEU. SARI suggests that multitask translations are simpler than baseline translations, and their resulting complexity correlates better with reference grade levels according to PCC.

The two pipeline models use the same MT system, therefore the difference between them comes from text simplification: using English simplification (first pipeline) outperforms Spanish simplification (second pipeline) according to BLEU and PCC, but not SARI. This can be explained by the smaller amount of Spanish simplification training data, which yields a model that generalizes poorly.

The ``All tasks'' model highlights the strengths of the multi-task approach: combining training samples from many tasks yields improvements over the ``Translate and Simplify'' multi-task model which is trained on the exact same data as the pipelines. However, even without additional training data, the multitask ``Translate and Simpifly'' model improves over baselines mainly by simplifying the output more, which suggests that the simplification component of the multitask model benefits from the additional MT training data.

Qualitative analysis suggests that the multi-task model is capable of distinguishing among different grade levels and the simplification operations performed for different grade levels are gradual. Table ~\ref{operations} illustrates simplification operations observed for a fixed grade 12 Spanish input into English with target grade levels ranging from 9 to 3. When translating to a nearby grade level, for example 9, the model roughly translates the entire input. For lower grade levels such as 7 and 5, lexical simplification and sentence splitting is observed. For the simplest grade level, the model deletes additional content. More examples are provided in the Appendix ( Table~ \ref{operations_more}).

\begin{table}[ht]
\centering
\begin{tabular}{lp{0.75cm}p{0.75cm}p{0.75cm}}
  \toprule
Complexity cont. MT & BLEU & SARI & PCC\\
 \midrule
\textit{Pipeline Baselines}\\
~Translate then Simplify & 21.98 & 30.4 & 0.436  \\
~Simplify then Translate & 17.09 & 37.4 & 0.275  \\
\textit{Multitask Models}\\
~Translate and Simplify & 22.51 & 44.8 & 0.572\\
~All Tasks & \textbf{22.75} & \textbf{45.0}  &  \textbf{0.608} \\
\bottomrule
\end{tabular}
\caption{Compared to pipeline models, multitask models produce complexity controlled translations that better match human references (BLEU), that are simpler (SARI), and whose resulting complexity correlates better with the target grade level (PCC). Pipeline models are trained on Newsela Simplification data and MT parallel data from Newsela and OPUS. ``Translate and Simplify'' uses the exact same data in a multi-task model. The ``All tasks'' model uses all data available, including Newsla MT+Simplify examples.\label{tab:mainresult} 
} 
\end{table}

\ignore{
The multitask model yields better bleu\mc{BLEU is an acronym and should always be capitalized} scores to the  pipeline approaches both for the task of Monolingual (En-En) Text Simplification and the MT+simplify task(see Table: \ref{multitaskvpipelineesen}). We also observe a significant boost in the SARI score for the task of MT+simplify which suggests that the model is able to generate simpler text but the target supervision is not sufficient to produce simplified output at a specific grade level as suggested by ARI adjacency accuracy. 

The significant drop in the performance of the pipeline model that performs simplification first(Row 2) can be attributed to the less number of available examples for the training of Spanish simplification model.
When preserving complexity in translation (MT+preserve task), the pipeline model performs better based on BLEU and ARI adjacency accuracy. This can be explained by the fact that the baseline for this task only uses the translation component of the pipeline, while the multi-task model is exposed to samples of more diverse complexity levels in training which damage its ability to preserve complexity. We don't quote SARI score for MT+preserve task for the baseline models as SARI score(Section: \ref{SARI}) compares the system output with both the input and references. Given our modified definition of computing SARI score for complexity controlled MT, the translated input and the system output are the same.
}

 \begin{table*}[ht]
\centering
\begin{tabular}{lp{14.5cm}}
  \toprule
12 & 	Ahora el museo Mauritshuis está por inaugurar una exposición dedicada a los autorretratos del siglo XVII, que destaca las similitudes y diferencias entre las fotos modernas y las obras de arte históricas. \\
 \midrule
9 & Now the museum Mauritois is launching an exhibition dedicated to the 18th century authoritations, highlighting the similarities and differences between modern photos and historical artworks.\\
7 & The museum is \textbf{now set to open} an exhibition dedicated to the 18th century authoritations, highlighting the similarities and differences between modern photos and historical artworks. \\
5 & The museum is now set to open an exhibit dedicated to the 18th century. \textbf{It highlights} the similarities and differences between modern photos and historical artworks. \\
3 &	The museum is now set to open an exhibit dedicated to the 18th century. It \textbf{shows} the similarities and differences between modern photos and art works.\\
\bottomrule
\end{tabular}
\caption{Example of multi-task model outputs when translating grade 12 Spanish into increasingly simpler English.\label{operations}} 
\end{table*}

\begin{table*}[ht]
\centering
\begin{tabular}{llllllllllll}
  \toprule
Adj.  &  & \multicolumn{10}{l}{Source Grade - Target Grade } \\
 level  &  Model & 1 & 2 & 3 & 4 & 5 & 6  & 7 & 8 & 9 & 10 \\
 \midrule
 1 & Pipeline & \textbf{0.593} & \textbf{0.629} & 0.594 & 0.556 & 0.524 & 0.493 & 0.472 & 0.457 & 0.448 & 0.444 \\
 & Multitask & 0.59 & 0.626 & 0.594 & \textbf{0.561} & \textbf{0.529} & \textbf{0.504} & \textbf{0.482} & \textbf{0.467} & \textbf{0.458} & \textbf{0.453} \\
2 & Pipeline & \textbf{0.717} & \textbf{0.759 }& \textbf{0.786} & 0.747 & 0.713 & 0.678 & 0.654 & 0.637 & 0.626 & 0.621 \\
& Multitask & 0.711 & 0.755 & 0.784 & \textbf{0.753} & \textbf{0.725} & \textbf{0.696} & \textbf{0.67} & \textbf{0.653} & \textbf{0.642} & \textbf{0.636} \\
\bottomrule
\end{tabular}
\caption{Adjacency ARI accuracy within grade level given by Adjacency level for the system output with respect to the target grade: Multitask model is able to better capture the target grade than the pipeline model when the difference between the source and the target grade is greater than 3. \label{gradediff}} 
\end{table*}

\section{Analysis}

\subsection{Output Grade Analysis}

We aim to better understand to what degree models simplify the input text: how often does the output complexity exactly matches that of the reference? Does this change depend on the distance between input and output complexity levels?  Table~\ref{gradediff} compiles Adjacency Accuracy scores \citep{HeilmanCollinsEskenazi2008}, which represent the percentage of sentences where the system output complexity is within $1$ or $2$ grades of the reference text. We derive the reading grade levels from ARI \cite{SenterSmith1967} and conduct this analysis for the best pipeline  (``Translate then Simplify'') and multitask models (``All Tasks''). These adjacency scores are broken down according to the distance between input and target grade levels.

When the source and target grades are close, roughly 60\% of system outputs that are within a $\pm1$ window of the correct grade level. The pipeline model matches the target grade slightly better than the multitask model. However, in the more difficult case where the difference between source and target grades is larger than three, the multitask model outperforms the pipeline. Increasing the adjacency window to $\pm2$ pushes the percentage of matches in the 70s. 

Overall these results show that multitask and pipeline models are able to translate and simplify, but that they do not yet fully succeed at precisely controlling the complexity of their output to match a specific target reading grade.

\subsection{Ablation Experiments}

Table~\ref{tab:tasks} shows the impact of different training data types on the multitask model using ablation experiments. OPUS improves BLEU and SARI performance across the board. However, using OPUS without any Newsela MT data (Row~4) hurts the correlation score, indicating the importance of in-domain MT data to control complexity.  The difference between the performance when using joint translation and simplification (MT+S) examples  (Row~2) vs. simplification only (S in Row~3) is small in terms of BLEU (+0.11) and PCC (0.012), indicating that the monolingual simplification dataset can provide simplification supervision when MT+Simplify data is unavailable. The overall best performance for the task is obtained by using all types of training examples.\footnote{A random sample of outputs from the best model configuration are provided in the Appendix (Table~ \ref{example_translations}).} 

\begin{table}[ht]
\centering
\tabcolsep=0.1cm
\begin{tabular}{cccccllll}
\toprule
\multicolumn{3}{c}{Newsela} & \phantom{a}  & OPUS & \phantom{a} & \multicolumn{3}{c}{Evaluation Metrics} \\
\cmidrule{1-3} \cmidrule{5-5} \cmidrule{7-9}
S & MT+S &  MT &  & MT & &  BLEU & SARI & PCC \\
\midrule
\cmark & \cmark & \cmark & & \cmark & & 22.75 & 45.0 & 0.608 \\
& \cmark & \cmark & & \cmark & & 22.62 & 44.5 & 0.584 \\
\cmark & & \cmark & & \cmark & & 22.51 & 44.8 & 0.572 \\
\cmark &  & & & \cmark  &  & 19.16 & 43.4 & 0.468 \\
\cmark & \cmark & \cmark & & & & 14.65 & 41.4 & 0.521\\
\bottomrule
\end{tabular}
\caption{Data ablation experiments showing the impact of different types of training examples on multi-task model. The OPUS parallel corpus is essential to good performance. Simplification data (S) can be used for simplification supervision when joint translation and simplification examples (MT+S) are unavailable.
\label{tab:tasks}} 
\end{table}

\subsection{Evaluation on Auxiliary Tasks}

In addition to complexity controlled MT, the multi-task model can be used to simplify English text, and to translate from Spanish-to-English without changing the complexity.  For completeness, we evaluate on these two auxiliary tasks.

Table~\ref{auxillary} summarizes the results: the multitask model slightly outperforms a dedicated simplification model on English simplification, showing the benefits of the additional training data from other tasks. By contrast, on the resource-rich MT task, the standalone translation system performs better. This can be explained by the fact that the standalone system is only responsible for text translation, while the multi-task model is exposed to samples of more diverse complexity levels during training which damage its ability to preserve complexity.

\addtolength{\tabcolsep}{-1.75pt}  
\begin{table}[ht]
\centering
\begin{tabular}{llll}
  \toprule
Task & BLEU & SARI & PCC  \\
 \midrule
\textit{English Simplification}\\
~Simplify & 55.76 & 41.7 & 0.736 \\
~Translate and Simplify & \textbf{56.47} & 41.3 & 0.730 \\
~All Tasks & 56.05 & \textbf{42.1}  & 0.736  \\

\textit{Machine Translation}\\
~Translate & \textbf{29.09}  & - & \textbf{0.769} \\
~Translate and Simplify & 27.33 & - & 0.647 \\
~All Tasks &  27.63 & -  & 0.658 \\

\bottomrule
\end{tabular}
\caption{Evaluation on auxiliary tasks: Multitask models trained on both the translation and simplification dataset improves the performance for the task of English Simplification. \label{auxillary}} 
\end{table}
\addtolength{\tabcolsep}{1.75pt}

\ignore{
\begin{table*}[ht]
\centering
\begin{tabular}{|p{3 cm}|l|l|l|l|l|l|l|l|}
  \hline
Model & \multicolumn{3}{l|}{En-En Simplification} & \multicolumn{2}{l|}{Es-En MT+preserve} & \multicolumn{3}{l|}{Es-En MT+simplify } \\
 \cline{2-9}
 & Bleu & Adj. Acc. & SARI & Bleu & Adj. Acc. & Bleu & Adj. Acc. & SARI  \\
 \hline
Translation followed by simplification & 55.76 & 34.9  & 41.7 & \textbf{29.09} & \textbf{37.3} & 21.98 &24.73 &  30.4\\
\hline
Simplification followed by translation & - & - &  - & 29.09 & 37.3 & 17.09 & 20.0 & 37.4 \\
\hline
Multitask & \textbf{56.47} & 33.5 & 41.3 & 27.33 & 26.1 & 22.51 & 21.0 & 44.8
\\
\hline
Multitask (all)  & 56.05 & \textbf{34.2} & \textbf{42.1} &  27.63 & 27.2 & \textbf{22.75} & \textbf{25.74}   & \textbf{45} \\
\hline
\end{tabular}
\caption{Multitask model trained on Newsela Simplification En-En, Newsela Es-En MT+Preserve (Translation only) and Opus corpus using human annotated grade as supervision performs better based on BLEU and SARI than pipeline model trained on the same datasets independently for simplification and translation.\label{multitaskvpipelineesen}\mc{TODO: clean up this Table to only include simplification and translation and to use the same style as other tables}} 
\end{table*}
}

\subsection{Provenance of Reading Grade Level}

Our models control complexity using the gold reading grade level assigned by professional Newsela editors. We investigate the impact of replacing these gold labels by automatic predictions from the ARI metric. ARI can be computed for any English segment, including for MT parallel corpora that are not annotated for complexity.

Table~\ref{tab:grade}  shows  that ARI  provides  an  adequate  substitute  for manually annotated reading grade levels, as BLEU and SARI score remain close when Newsela reading grade levels are replaced by ARI-based tags.  However, annotating all data with ARI grades, including the OPUS parallel corpus, hurts BLEU. We attribute this result to the differences in length and number  of sentences per segment in OPUS vs. Newsela (Table \ref{tab:stats1}): segments of vastly different lengths can have the same ARI score (Equation \ref{ariequation}), thus confusing the multitask model.

\begin{table}[t]
\centering
\begin{tabular}{llll}
  \toprule
Complexity cont. MT & BLEU & SARI & PCC \\
 \midrule
Newsela reading grade & 22.51 & 44.80 & 0.572\\
ARI on Newsela data & 22.26 & \textbf{45.12} & \textbf{0.581}  \\
ARI on all data & 20.91 & 44.75 & 0.577\\
\bottomrule
\end{tabular}
\caption{Complexity controlled MT with automatic vs. manual reading grade level tags: ARI  provides an  adequate  substitute for manually assigned  grade levels. 
\label{tab:grade}} 
\end{table}

\section{Conclusion}

We introduce a new task that aims to control complexity in machine translation output, as a proxy for producing translations targeted at audiences with different reading proficiency levels. We construct a Spanish-English dataset drawn from the Newsela corpus for training and evaluation, and adopt a sequence-to-sequence model trained in a multitask fashion.

We show that the multitask model improves performance over translation and simplification pipelines, according to both machine translation and simplification metrics. The reading grade level of the multi-task outputs correlate better with target grade levels than with pipeline outputs. Analysis shows that these benefits come from their ability to combine larger training data from different tasks.
Manual inspection also shows that the multi-task model successfully produces different translations for increasingly lower grades given the same Spanish input. 

However, even when simplifying translations, multitask models are not yet able to exactly match the desired complexity level, and the gap between the complexity achieved and the target complexity increases with the amount of simplification required.  
Our datasets and models thus provide a foundation to investigate strategies for a tighter control on output complexity in future work. 




\bibliography{emnlp-ijcnlp-2019}
\bibliographystyle{acl_natbib}

\appendix
\newpage

\onecolumn

\section{Supplemental Material}
\label{sec:supplemental}

Table \ref{tab:stats2} and \ref{tab:stats3} provides the statistics of grade pair distribution in the Newsela English and Newsela Spanish-English dataset. 

\begin{table*}[ht]
\centering
\begin{tabular}{l|lllllllll}
  \toprule
Src / Tgt & 2 & 3 & 4 & 5 & 6 & 7 & 8 & 9 & 10\\
 \midrule
3 & 2652 & 0 & 0 & 0 & 0 & 0 & 0 & 0 & 0 \\

4 & 4984 & 8212 & 0 & 0 & 0 & 0 & 0 & 0 & 0 \\

5 & 2287 & 19589 & 23775 & 0 & 0 & 0 & 0 & 0 & 0 \\

6 & 1914 & 7625 & 21022 & 21380 & 0 & 0 & 0 & 0 & 0 \\

7 & 608 & 8897 & 14249 & 33466 & 10944 & 0 & 0 & 0 & 0 \\

8 & 623 & 3710 & 13267 & 17347 & 22745 & 12006 & 0 & 0 & 0 \\

9 & 130 & 5058 & 5031 & 19834 & 4684 & 30929 & 2144 & 0 & 0 \\

10 & 6 & 40 & 224 & 320 & 382 & 289 & 400 & 142 & 0 \\

11 & 0 & 0 & 15 & 19 & 11 & 16 & 28 & 0 & 0 \\

12 & 1069 & 6818 & 18430 & 34232 & 28532 & 41561 & 29836 & 31327 & 97 \\
\bottomrule
\end{tabular}
\caption{Number of text segments per grade level pair in our Newsela English Corpus
}\label{tab:stats2}
\end{table*}

\begin{table*}[ht]
\centering
\begin{tabular}{l|lllllllll}
  \toprule
Src / Tgt & 2 & 3 & 4 & 5 & 6 & 7 & 8 & 9 & 10\\
 \midrule

3 & 293 & 0 & 0 & 0 & 0 & 0 & 0 & 0 & 0 \\

4 & 670 & 1305 & 0 & 0 & 0 & 0 & 0 & 0 & 0 \\

5 & 251 & 3383 & 1957 & 0 & 0 & 0 & 0 & 0 & 0 \\

6 & 223 & 1124 & 2090 & 2833 & 0 & 0 & 0 & 0 & 0 \\

7 & 60 & 1249 & 926 & 4986 & 1244 & 0 & 0 & 0 & 0 \\

8 & 96 & 548 & 1016 & 1804 & 3705 & 1221 & 0 & 0 & 0 \\

9 & 16 & 717 & 211 & 3074 & 189 & 6135 & 263 & 0 & 0 \\

10 & 0 & 3 & 5 & 15 & 26 & 1 & 46 & 0 & 0 \\

12 & 189 & 1288 & 1902 & 5312 & 4708 & 7796 & 4995 & 7077 & 30 \\
\bottomrule
\end{tabular}
\caption{Number of text segments per grade level pair in our Newsela English-Spanish Corpus
}\label{tab:stats3}
\end{table*}

\begin{table*}[ht]
\centering
\begin{tabular}{llll}
 \toprule
Model & Bleu & SARI & Flesch \\
    \midrule
\textit{Results copied from \newcite{ScartonSpecia2018}}\\
  ~seq2seq w/ side-constraint  & 62.91 & 41.01 & 82.91  \\
\textit{Reimplementation evaluated on our Newsela download}\\ 
~seq2seq w/ side-constraint & 58.61 & 39.81  & 70.44\\
~seq2seq w/ side-constraint + BPE  & 61.87	& 52.78 & 66.98\\
\bottomrule
\end{tabular}
\caption{Comparison with previously published results on Newsela English text simplification. Our implementation yields BLEU and SARI score that are close to those reported in \newcite{ScartonSpecia2018}. The difference in Flesch score can be attributed to changes in the number and complexity of articles available in newsela at the time the datasets were extracted.}\label{tab:resultsmono} 
\end{table*}


 \begin{table*}[ht]
\centering
\begin{tabular}{lp{15cm}}
  \toprule
12 &	Se estima que 75 personas han expresado interés en alojar al menos a un solicitante de asilo, dijo Cronk. Algunas de estas personas viven en las principales áreas metropolitanas de California y Nueva York. Otros son de zonas remotas y rurales de Montana y Dakota del Norte. \\
 \midrule
8 & An estimated 75 people have expressed interest in hosting at least one asylum-seeker said Cronk. Some of these people live in major California and New York area. Others are from remote and rural areas of Montana and North Dakota. \\
6 & An estimated 75 people have expressed interest in hosting at least one asylum-seeker said Cronk. Some of these people live in major California and New York area. Others are from remote and rural areas of Montana and North Dakota. \\
4 &  An estimated 75 people have expressed interest in hosting at least one asylum-seeker said Cronk. Some of these people live in the \textbf{main} areas of California and New York. Others are from remote and rural areas of Montana and North Dakota.\\
2 &	\textbf{Many people live in the} United States and New York City. Some are from remote areas of Montana and North Dakota.\\
\midrule
12 & El gobierno federal realizó un contrato con el centro de detención juvenil en Vicennes, Indiana, desde el año 2004 hasta el año 2010, para que alojara a aquellos niños inmigrantes considerados como los más peligrosos.\\
\midrule
9 & The federal government conducted a contract with the youth detention center at Vicennes, Indiana, since 2004 to 2010 to host those immigrant children considered as the most dangerous ones. \\
7 & The federal government conducted a contract with the youth detention center at Vicennes, Indiana, since 2004 to 2010, \textbf{so that they hosted} those immigrant children considered as the most dangerous ones.\\
5 & The federal government \textbf{made} a contract with the youth detention center in Vice-Year. \textbf{It is in the} United States since 2004 to 2010.\\
3 & The government made a deal with the youth detention center in Vice-Year. It is in the United States since 2004 to 2010.\textbf{ It was to host} those immigrant children as the most dangerous in 2010.\\
\midrule
12 & Poco después de que el dron despegó durante la prueba de Verizon en Cape May, una "aeronave de seguimiento" salió tras él, para garantizar que el dron pudiera evadir otros aviones en caso de volar dentro de un espacio aéreo designado.\\
\midrule
9 & Shortly after the drone took off from Verizon in Cape May, a "unmanned aircraft following" came out after him, to ensure that the drone could evade other planes in case of flying inside a designated airspace. \\
6 & Shortly after the drone took off in Cape May, a "unmanned aircraft force" came out \textbf{behind} him. \textbf{They could} ensure that the drone could evade other planes in case of flying inside a designated airspace. \\
2 & \textbf{Not long} after the drone took off from Verizon in Cape May, a "unmanned aircraft help" came out after him. \textbf{It is to make sure} the drone could evade other planes in case of flying inside a air space. \\
\bottomrule
\end{tabular}
\caption{ Examples of simplification operations observed when simplifying from a higher Grade level into different lower grade levels using the multitask model (All tasks).\label{operations_more}} 
\end{table*}

\begin{table}[ht]
\centering
\begin{tabular}{llp{4cm}p{4cm}p{4cm}}
  \toprule
Src\_G & Tgt\_G & Source & Target & Predicted\\
\midrule
8 & 5 & Los partidarios de la reserva dicen que marca un precedente para muchos países que trabajan conjuntamente para proteger una gran porción del océano que no se halla dentro de la jurisdicción específica de ningún país. &  Supporters of the reserve say it sets an example for multiple countries working together to protect a large area of ocean. The area is not controlled by any single nation. & Supporters of the reserve say that marks a precedent for many countries that work together to protect a large portion of the ocean. \\
\midrule

5 & 3 & Los padres no respondieron bien al escuchar que sus hijas estaban surfeando porque les preocupaba el peligro y la reputación de las niñas.&The parents did not react well to their daughters surfing. They worried about the danger and what other people would think. & Parents did not want to hear that their daughters were surfing their daughters because they were worried about the danger and their reputation. \\

\midrule
12 & 5 & Según dijo el actor Eds Eramiha, esta escena habría sido difícil de imaginar hace tan solo dos décadas, cuando el idioma te reo maorí era ampliamente considerado una lengua moribunda que no valía la pena enseñar. & Eds Eramiha is an actor in the play. He says it would have been difficult to imagine this scene just 20 years ago. Back then, te reo Maori was widely regarded as a dying language not worth teaching. & Actress Eds Eramiha said this scene would have been hard to imagine only two decades ago. The language was widely viewed as a dying tongue that would not be worth teaching.
\\
\midrule
5 & 4 & Dos astrónomos han planteado una fascinante pregunta: ¿Pueden las lunas tener lunas? La interesante respuesta es que sí, una luna puede tener su propia luna. & Two astronomers asked a question: Can moons have moons? The interesting answer is yes, a moon can have its own moon. & Two astronomers have asked a fascinating question: Can Mondays have Monday? The interesting answer is that yes, a moon can have its own. \\
\midrule
9 & 3 & Como hijo de un sastre y modisto de pocos ingresos, consiguió trabajo en el taller de un sastre en Brooklyn a la edad de 15 años, terminó su secundaria estudiando por las noches y luego se dispuso a triunfar en la vida. & Campos is the son of a tailor and a poor dressmaker. He found work in a Brooklyn clothing shop at age 15  and  He finished high school at night. Then he set out to succeed. & The son of a saga and a modest age of a few incomes, he got work at the age of 15 years old. He graduated from the evenings and then turned into life in life. \\
\bottomrule
\end{tabular}
\caption{Example translations produced by our best multitask model. Refer Table \ref{tab:tasks} (Row 1). \label{example_translations}} 
\end{table}

\end{document}